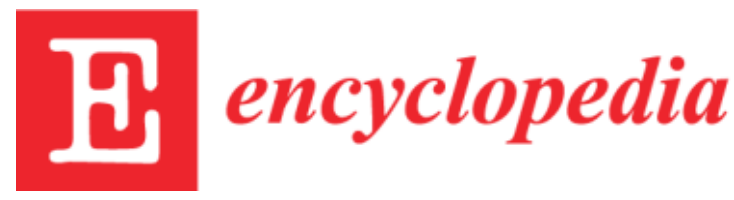
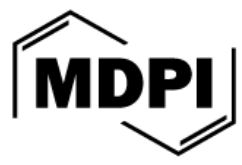

Review

# Deepfakes and Synthetic Media: Generation, Detection, and Governance

**Alexandros Gazis [1,2,*], Efstathios Karypidis [3], Kleanthi Santamouri [4,5], Theodoros Vavouras [6,7], Nikos E. Mastorakis [8,9] and Stylianos Pappas [9]**

[1] Edinburgh Business School, Heriot-Watt University, Edinburgh EH14 4AS, UK
[2] Department of Electrical and Computer Engineering, School of Engineering, Democritus University of Thrace, 67100 Xanthi, Greece
[3] School of Engineering, National and Technical University of Athens, 15780 Athens, Greece; stathiskaripidis@gmail.com
[4] Private Primary Education, Athens, Greece; a.santamouri@salvezza.eu
[5] Salvezza Energy Systems Ltd., Sirakouson 4, Paphos 8016, Greece
[6] Department of Humanities, School of Humanities, Hellenic Open University, 26335 Patras, Greece; vavouras.theodoros@ac.eap.gr
[7] Department of Philosophy, School of Italian Language and Literature, Aristotle University of Thessaloniki, 54124 Thessaloniki, Greece
[8] English Language Faculty of Engineering, Technical University of Sofia, 1756 Sofia, Bulgaria; mastor@tu-sofia.bg
[9] Electrical Engineering and Computer Science, Hellenic Naval Academy, Terma Chatzikyriakou, 18539 Piraeus, Greece; steliospappas@teemail.gr
* Correspondence: agazis@teemail.gr

## Abstract

Deepfakes, synthetic audiovisual content produced by deep generative models, have escalated into a critical threat across civilian and military domains, enabling identity fraud, disinformation campaigns, and evidence fabrication. In high-stakes environments, ranging from journalism and finance to healthcare and legal contexts, the consequences extend to severe misinformation, market manipulation, identity fraud, and the erosion of institutional trust. This entry explores how modern visual intelligence and computer-vision techniques are used to detect deepfakes. It outlines key deepfake generation models, such as GANs, autoencoders, neural rendering, and diffusion systems, while also explaining how adversarial methods enhance realism and challenge existing detectors. The overview highlights visual artifacts, digital patterns, and physiological cues commonly leveraged in detection and reviews major CNN, transformer, and frequency-based approaches. It also summarizes evaluation practices and the difficulty of achieving strong generalization. Finally, it identifies emerging directions, including modern intelligence techniques for civilian and military content verification. This survey covers generation architectures (GANs, latent diffusion, neural rendering, video synthesis), the spatial, temporal, frequency-domain, and physiological artifacts they produce, and the detector families that exploit them. We examine evaluation benchmarks and protocols, highlighting cross-generator generalization as the field's central open challenge. Beyond detection, we discuss cryptographic provenance standards, watermarking, and regulatory frameworks (EU AI Act, DSA, GDPR). We conclude that effective deepfake governance requires defense in depth integrating forensic detection, verifiable provenance, and institutional accountability.

## 1. Definition, Classification and Why Deepfake Is Important

The term deepfake commonly refers to synthetic or heavily manipulated media, most notably images, videos, and speech, whose realism is enabled by modern machine learning, particularly deep generative models. In practice, the boundary between "deepfakes" and other manipulated media is increasingly blurred: contemporary pipelines combine generative components (e.g., diffusion or Generative Adversarial Network (GAN)-based synthesis), classical editing (compositing, retouching), and most notably, post-processing optimized for distribution environments (compression, resizing, platform-specific transcoding).

A handful of practical illustrative examples may be used to clarify the above. Specifically, everyday edits can change the look of a media, without changing what it depicts. This can be done, for example, via color grading, cropping, sharpening, compression, contrast, and other edits. Furthermore, as one travels further down the scale, there exists a wide range of manipulations, including object removal, face retouching, background replacement, or AI-assisted inpainting, which alter aspects of a photo or video, while usually keeping the event recognizable. For instance, a news picture might show a digitally slimmed crowd or erased banner, which alters the impression of an occurrence without creating the presence or words of a person. Similarly, at the end of the spectrum, techniques like face swapping, lip-sync manipulation, voice cloning, or even complete text-to-video generation can falsely assign fabricated words, deeds, or experiences to someone. As such, the ability to define what deepfake is depends on whether media elements are not—yet—a simple visual enhancement, but an alteration of who someone appears to be. Lastly, the modification may also involve what the media may be used to prove or what they seem to have believed or done; thus, the variations and possibilities are limitless.

This issue is important because it can affect both everyday civilian communication and military or security-related applications. As a result, an operational definition for researchers and practitioners is functional. A deepfake refers to media whose perceptual plausibility is sufficiently high to create a credible false impression about who did what, when, or where, and whose creation is materially facilitated by AI-based generative or reenactment methods [1–3]. This is of great importance, across diverse sectors; in journalism, it can skew public perception, in finance and healthcare it can lead to devastating fraud or medical misinformation, and in legal contexts, it threatens the integrity of audiovisual evidence. For example, many deepfake systems learn latent representations of identity, pose, expression, voice, or scene structure, and then recombine them to produce new audiovisual output. As a result, this enables systems to create the illusion that a target person is speaking, moving, or acting in a way that did not occur.

In Figure 1, we illustrate this functional taxonomy and how it separates deepfakes into three complementary dimensions: modality, manipulation intent, and generation regime. As such, based on the functional definition, one can classify deepfakes into the following broad categories based on their usage, properties, and functionality [4]:

1. Modality deepfakes
   a. Visual: face swaps, reenactment, lip-sync, attribute editing (age, expression), full-body synthesis, scene relighting.

b. Audio: voice cloning (speaker identity conversion), speech-to-speech conversion, text-to-speech impersonation.
   c. Multimodal: synchronized audio–video generation, avatar systems, "talking head" models with cloned voice.
2. Manipulation intent deepfakes
   - Identity substitution (impersonation, fraud, non-consensual sexual content).
   - Event fabrication (false evidence, fake statements, fake presence).
   - Contextual distortion (true footage reframed via synthetic overlays, selective edits, or deceptive narration).
3. Generation regime deepfakes
   a. Closed-world generation (trained on a specific target identity).
   b. Open-world generation (foundation models enabling broad, low-friction synthesis).

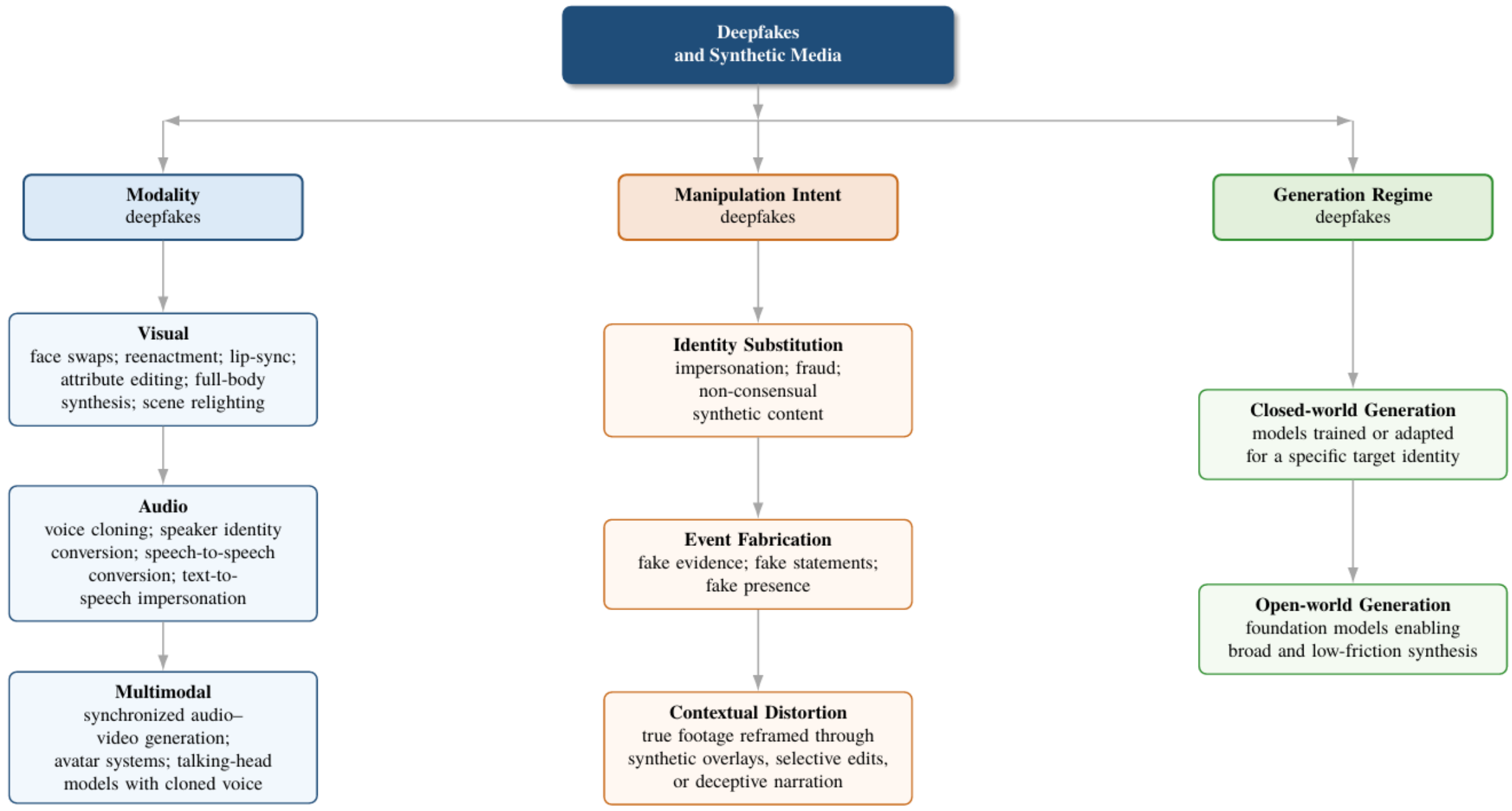

**Figure 1.** Functional taxonomy of deepfakes and synthetic media according to modality, manipulation intent, and generation regime.

The above taxonomy, even though generic and abstract, showcases real value as it suggests that detection and governance depend on which deepfake type is targeted. This distinction is also relevant for military risk assessment, where impersonation, event fabrication, and contextual distortion may produce different operational consequences. Lastly, it is noted that many failures of detection methods in real-world deployments stem from treating "deepfake detection" as a single, uniform classification problem.

### *1.1. Why Are Deepfakes Important?*

Deepfakes are more than a computer-vision issue; they are a socio-technical risk shaped by actors, incentives, and distribution channels. Analytically, in a typical threat model, an adversary has either access to a target's public media (photos, interviews, livestreams, military briefings, or open-source operational footage), commodity tools to generate or edit content, or platforms to distribute at scale [5–7]. This means that the harm is amplified by platform dynamics (virality, recommender systems) and by asymmetric verification costs (it is cheaper to generate than to thoroughly authenticate) [8,9].

A practical implication is that risk management should be framed at the system level, covering creation, publishing, detection, and incident response, rather than relying on "a model that flags deepfakes" as a standalone solution. This is especially important for military and security organizations, where false positives and false negatives may influence operational decisions. Moreover, it aligns with governance-oriented frameworks that treat AI risk across the full lifecycle and operational context [10].

As such, deepfakes combine high realism with personalization [11]: they can be tailored to a specific individual and context, which increases plausibility and emotional impact. For example, personalization can be used to imitate corporate executives authorizing fraudulent financial transfers, healthcare professionals giving false medical advice, or public figures in critical news broadcasts. As such, even when a given deepfake is debunked, the broader informational ecosystem can still be damaged via two mechanisms:

- Evidentiary erosion: Over time, authentic recordings become easier to dismiss as fake (e.g., someone might state: "this is not genuine, it could be AI").
- The liar's dividend: Public figures can strategically exploit uncertainty about synthetic media to evade accountability, deny authentic evidence, or muddy public understanding [12,13].

Therefore, the deepfake problem is not only "fake media exists" [1], but "the credibility of media as evidence is destabilized". This is important both in terms of journalism and informed civil decisions as well as in regard to military communication, where it can weaken the trust in genuine battlefield footage, official announcements, or intelligence material.

### *1.2. Why Is Detection Necessary but Not Always Sufficient?*

State-of-the-art detection has progressed rapidly, but deepfake defense is an arms race: generation improves, artifacts shift, and distribution transforms signals (compression, re-encoding) [14,15]. Specifically, deepfake detection attempts to figure out if the statistical patterns of an image, video, or audio signal are more compatible with natural capture or synthetic generation. As modern generators limit obvious visual artifacts, a robust detection framework must combine spatial, temporal, frequency-domain, physiological, and provenance-based evidence. Deep convolutional neural networks have consistently outperformed classical feature-based approaches for visual recognition tasks [16,17], motivating their central role in forensic detection. For military users, this defense-in-depth model should also include forensic review, provenance checks, and command-level verification procedures. As such, for this reason, current best practice is defense in depth, combining [18]:

1. Forensic detection (model-based classifiers, artifact analysis, physiological/temporal cues) [1–3,19].
2. Provenance and authenticity mechanisms (cryptographic manifests, signed metadata, content credentials, watermark recovery workflows), which aim to answer common questions such as: "where did this come from?" rather than the usual generic: "does it look fake?" [20,21].
3. Policy and compliance controls, including transparency duties for certain AI outputs and platform obligations for risk mitigation and accountability. In the European Union context, deepfakes intersect directly with:
    a. The AI Act (risk-based obligations; transparency requirements for certain synthetic or manipulated content contexts) [22];
    b. The Digital Services Act (systemic risk management and transparency duties for platforms) [23];

c. General Data Protection Regulation (GDPR) (identity, biometric data, lawful basis, data subject rights) [24,25].

These instruments do not replace technical measures, but they shape operational requirements, documentation, and response procedures.

### 1.3. Human Factors and Cognitive Vulnerabilities

Most of the deepfake impact beyond technical generation and algorithmic detection has deep social roots within human realms [7,11]. Specifically, over the years, it has been ingrained in our evolution to trust audiovisual evidence [13]. More specifically, to see and to hear traditionally ultimately meant to believe. As such, the evasion of analytical scrutiny by high-fidelity synthetic media takes place when emotional responses which outsmart cognitive defenses deployed by humans become activated [11,15]. This is important if we consider that confirmation bias has a marked impact on user perception [15]. Users, for example, are more likely to willingly accept, believe, and share misinformation that corresponds with their preexisting views, political affiliation, or emotional response to an issue [15].

As such, it is important to notice that a cognitive bias exists, causing one to misperceive events objectively [13,15]. This can be interpreted by highly personalized or emotionally manipulative deepfakes which influence public perception and trust, even when they do not exhibit visual artifacts. Therefore, we must deploy systemic strategies aiming at reducing cognitive vulnerabilities through advanced media literacy, even as robust technical detection invention is important [9,20].

## 2. Creating Deepfakes: Types of Fakes, Standard Pipelines, and Generative Models

On a technical level, deepfakes are generated by generative modeling and reenactment/editing systems, which transform audiovisual information to produce a more convincing but inauthentic result. This is equally relevant to civilian media and military information operations. In practice, implementations combine several methods such as synthesis models (GANs/diffusion/neural rendering), classic processing stages (compositing/blending), and, most recently, post-processing adapted to platform environments (compression, re-encoding, scaling). As such, a more in-depth technical understanding of "how deepfakes are produced" is necessary so one can accurately map the expected artifacts, generalization failures, and appropriate evaluation protocols.

### 2.1. Common Deepfake Types

Based on the above, the most common functional typologies, i.e., types of deepfakes that are detected in industry and academia alike, are the following:

- Face swapping: The target's face is replaced with the source's face to preserve the pose, lighting, and background. Historically, this was based on GAN families and evolved into high-fidelity models (e.g., the StyleGAN line) [26–28]. Such techniques are especially problematic for sensitive communications, including breaking news journalism, legal evidence submissions, and crisis management broadcasts.
- Face reenactment (expression/motion reenactment): The target's identity is preserved, but the expression is "driven" by a source (video/live stream), often with 3D parameterization and realistic rendering [29,30].
- Lip-sync/talking head editing: Primarily the mouth region is modified to match a new audio signal; Wav2Lip is a key benchmark for synchronization under "uncontrolled" conditions [31,32].

- Audio deepfakes (voice cloning/conversion): Speech is generated or transformed to mimic a specific speaker (voice conversion/text-to-speech), often as part of a full audiovisual deepfake [33–36]. For example, in military contexts, this may support voice-based impersonation or false command messages.
- Text-to-video/full scene synthesis: Video sequences (and not just "doctored" faces) are generated using diffusion-based video generation and text-to-video approaches that expand the threat from "evidence tampering" to "event fabrication" [37,38]. This is crucial as it creates a risk of fabricated military events, scenes, predictions, or operational incidents [39].

### *2.2. Typical Production Pipeline*

Although the software tools and methods used vary, most production workflows typically share some common stages [40]:

1. Data collection/selection: sufficient variety of poses, expressions, and lighting (especially for identity-specific models).
    - This stage is technically important as the model learns to identify specific variations from repeated examples under different poses, illumination conditions, facial expressions, and camera qualities. Limited or biased training material often leads to failures under unseen angles, lighting, or movements, which later become detectable artifacts.
2. Localization/normalization: face detection, landmarks, alignment, cropping, and photometric normalization.
    - The unconstrained visual input of this stage is preprocessed to yield a normalized facial representation. In this stage, the system focuses on localizing the face, estimating landmarks, aligning the facial region, and reducing background variation. Nonetheless, these methods might also result in geometric warping, disruptions at boundaries, or improper blending that forensic detectors can use.
3. Model training or adaptation: either general-purpose (foundation-style) or tailored to a specific individual/target.
    - At this point, the model will learn to map representations of the input data, e.g., facial landmarks, identity embeddings, motion parameters, and acoustic features, to realistic synthetic output. In identity-specific systems, adaptation or fine-tuning enables the model to reproduce the appearance, voice, or movement patterns of a particular target more convincingly at a cost of overfitting to the material available in training.
4. Inference and temporal consistency: especially in video, temporal consistency is crucial for perceptual plausibility.
    - At this stage, when a trained model is evaluated, it creates altered frames or audio segments from what it has learned, typically within a framework of the target pose, expression, or speech signal. Temporal consistency in video deepfakes is challenging in practice because each frame must be consistent with the previous and next one in terms of facial geometry, illumination, lip motion, eye movement, and background continuity.
5. Post-processing: blending, color matching, denoising/oversampling, and final re-encoding, which often "hides" obvious traces and shifts detection to more subtle statistical cues.
    - The last stage of the post-processing aims to ensure the produced region appears visually compatible with the rest of the image or video. This includes boundary smoothing, color difference correction, noise reduction, and adjustment to the

platform's encoding format. While these operations can mask obvious marks, they may also introduce other forensic signs, such as abnormal compression patterns, spectral irregularities, or more subtle inconsistencies between the manipulated region and the original background.

The above-mentioned production and response logic is summarized in Figure 2, which connects the technical generation pipeline with the subsequent layers of forensic detection, provenance verification, and governance-oriented response.

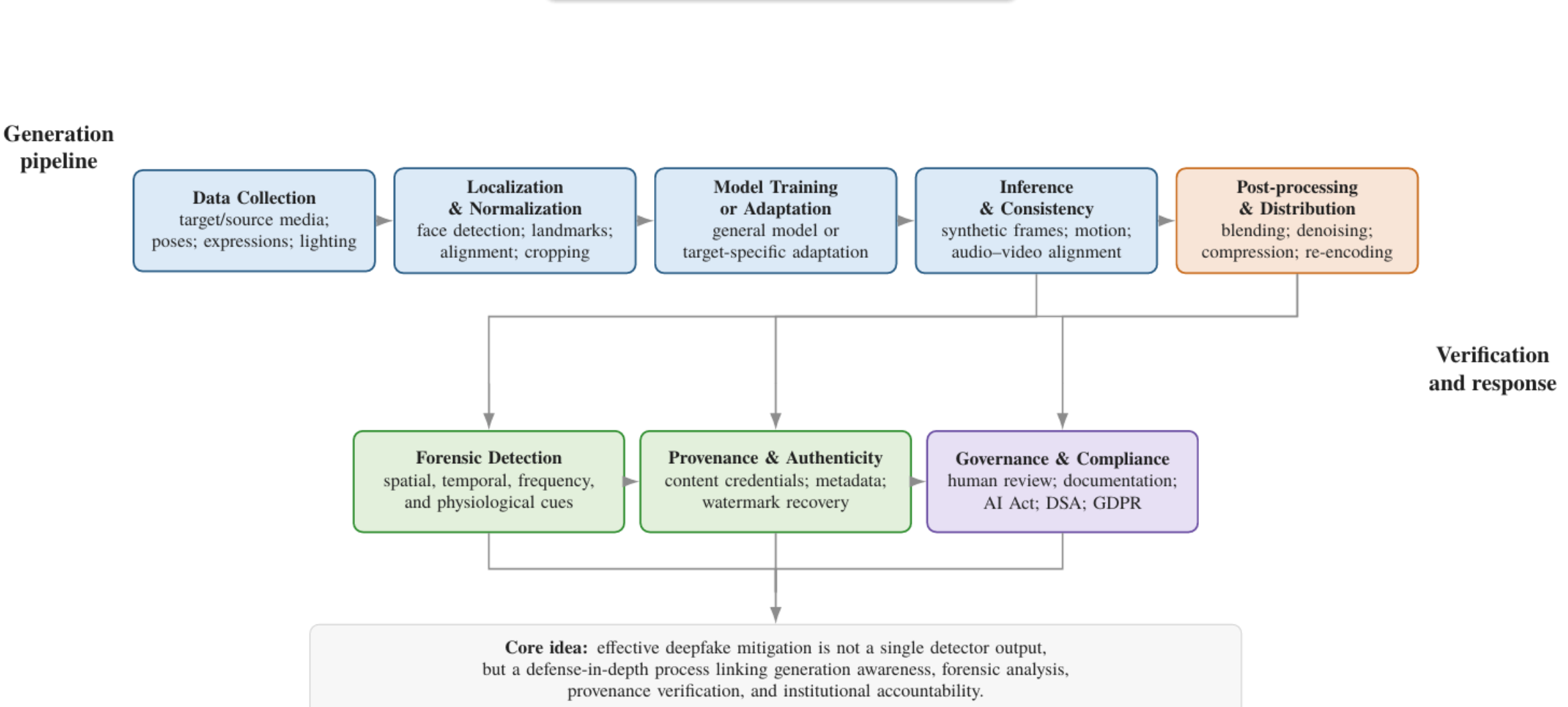

**Figure 2.** End-to-end deepfake lifecycle from data collection and generative processing to post-processing, forensic detection, provenance verification, and governance response.

### *2.3. Major Families of Generative Models*

We categorize deepfake generative models into four main families depending on the technical principles and output modality. The first category, focused on Face Image Generation, deals with the synthesis of photorealistic faces, image-to-image translation, and attribute editing, primarily using GAN-based architectures. The second family relates to 3D reenactment, neural rendering, and latent diffusion models, where facial motion /geometry/latent representations are utilized to synthesize manipulated content. The third set of families concerns image-to-video and video-generation methods through 3D-aware rendering. The fourth family concentrates on audiovisual deepfakes, specifically speech synthesis, voice cloning, and lip-sync generation. This section's goal is to describe these four families and their technical characteristics, as well as their implications for detection and assessment.

As for the first family of GANs, this occurred when GANs initially introduced the competitive generator–discriminator framework and laid the foundation for realistic image synthesis [26]. In this structure, the generator attempts to produce realistic samples while the discriminator tries to determine which samples are generated and which are real; these two networks are trained together through a competition process. This mechanism improves the texture of the face, the consistency of the lighting, and the realism of the identity. However, this may give rise to generator-specific statistical fingerprints. Analytically, the evolution toward high-fidelity architectures, such as [27], drastically increased texture detail and the stability of the generative process as GANs started to have a direct consequence on "visual plausibility" no longer being a sufficient criterion for authenticity. Moreover, unpaired image translation models, such as CycleGAN and StarGAN, have enabled feature/domain transformations without paired data and are used as building blocks for targeted modifications (e.g., expression, style, facial features) [41–43]. In such scenarios, the "identity" may remain intact, while semantically critical properties are altered.

Secondly, as for the family of models related to 3D reenactment, neural rendering, and latent diffusion models, the reenactment line was enhanced by 3D approaches: Face2Face demonstrated real-time expression transfer to red–green–blue (RGB) video through 3D parameterization and realistic rendering [29,30]. For instance, some familiar terms used in such methods include keywords such as head pose, expression coefficients, mouth movement, and illumination, which are often parameterized to represent facial motion; the modified face is then rendered back into the original video. This explains why reenactment deepfakes may look visually plausible at the level of individual frames, but still exhibit temporal or geometric artifacts across frames. Meanwhile, Deferred Neural Rendering introduced neural textures and a learnable rendering pipeline, reducing the need for "manual" blending techniques but introducing new statistical "signatures" into the output signal [44–46]. In addition, Denoising Diffusion Probabilistic Models (DDPMs) treat synthesis as a gradual denoising process and, in many cases, offer stable training and high-quality samples [14,39,47,48]. At this point, it is worth noticing that the main idea is that the model first learns how real data are progressively corrupted by noise, and then learns the reverse process, step-by-step reconstructing realistic data. Latent diffusion works in a compressed feature space for this denoising rather than the pixel space, reducing the computing cost and facilitating high-resolution synthesis. Denoising Diffusion Implicit Models (DDIMs) accelerated sampling using implicit procedures [49,50]. Practical scaling to high resolutions was achieved with latent diffusion, where generation occurs in a compressed latent space, drastically reducing computational cost [51,52]. For deepfakes, this broadens the scope from "face-centric" to generalized high-fidelity synthesis/editing.

The third family moves from static image manipulation to full video generation. Unlike single-image synthesis, video generation has to maintain spatio-temporal continuity; generated frames must stay consistent in identity, pose, lighting, background, and motion throughout the sequence. Video diffusion models extend diffusion architectures to handle the temporal dimension as part of the generative process [37,38]. Text-to-video approaches such as [38] show that video generation can draw on text–image data and unsupervised motion knowledge, cutting down the reliance on large text–video paired datasets. In practical terms, this shifts the threat from manipulated facial evidence to fully synthetic visual events—not just a fake person, but a plausible sequence of actions or scenes.

Neural Radiance Fields (NeRFs) [53] introduced continuous 3D scene representation from 2D images, making it possible to render new viewpoints with high consistency [54]. Neural Radiance Fields (NeRFs) are not a deepfake tool in itself, but 3D-aware compositing supports avatar generation and view-consistent rendering that reduces the geometric errors, odd angles, and inconsistent lighting that detection systems often rely on. This makes detection based purely on visible geometric mistakes harder to sustain.

Lastly, the fourth major family of generative deepfake models concerns audiovisual deepfakes, from voice cloning and lip-syncing to implications for detection and evaluation methods. Specifically, full impersonation often requires synchronizing voice and lip movements. Typically, audio deepfake systems leverage a separation between linguistic content and speaker identity by learning acoustic embeddings that provide timbre, rhythm, accent, and style. A vocoder converts those intermediate acoustic features into a waveform so that detection can examine the spectral smoothness, the phase behavior, and speaker-consistency cues. As such, in order to transfer a speaker's tone and style, recent studies such as [33] propose zero-shot voice style transfer using a lossy autoencoder.

Similarly, for high-fidelity waveform synthesis, vocoders such as [34] enable efficient and realistic performance, while other research such as [35] serves as a classic foundation for generative modeling of raw audio signals. On the visual side, Ref. [31] significantly improves speech-to-lip synchronization in real videos. As a result, these projects pave the way for detection that leverages cross-modal consistency (sound–lips) rather than just spatial artifacts.

As a conclusion, we may note that the evolution of generative models is shifting the focus from "artifact hunting" to multi-level defense. This is evident as newer generators reduce visible traces, post-processing and platform compression alter the signals that detectors "see," and most notably, the rise in text-to-video increases the need for out-of-distribution evaluation and for methods that do not rely exclusively on specific artifacts. For military applications, this also means that detection tools must be tested against both face-centric impersonation and broader synthetic event fabrication. Lastly, the proliferation of audiovisual deepfakes reinforces the importance of multimodal checks (audiovisual alignment) in detection.

### *2.4. Transition to Multimodal Models*

In our contemporary society, the early innovators of deepfakes primarily focused on face swapping technologies [28,40]. Therefore, while historically the generation of deepfakes has often relied on specific architectures for each task or target (for example: early autoencoders or GANs that are trained to only do one task (e.g.: swap a given face, clone a given voice)) [26,29,33], multimodal models aim to present a new solution: an algorithm that requires only one architecture for neural synthesis for any deepfake generation. As such, the current trend in generative models is leaning significantly towards large general-purpose multimodal foundation models [14,37]. This means that modern systems will operate as open-world generators that use huge datasets, transformers, and diffusion [47,51].

Moreover, modern generative AI systems now enable the simultaneous synthesis across several modalities by means of a zero-shot or few-shot approach without requiring exhaustive target-specific fine-tuning or paired datasets [38,41,52]. As a result, they can convert text prompts into a hyper-realistic video sequence (text to video) or create a sound in the visual [37,38]. This is of great importance as this change may result in a fundamental alteration in the threat model, whereby the technical barrier to entry as well as the scope of the synthetic media will be reduced greatly [17,20]. Based on this point, one can argue that the risk is not merely restricted to tight identity manipulation (e.g., face swapping) but includes open-world fabrication of complex events [38]. In light of this, it is necessary to highlight that detection methods gradually shift from the search for spatial artifacts to assessing multimodal semantic consistency [16,32].

## 3. Deepfake Characteristics and Detectable Properties

Building on the operational definition and taxonomy introduced in the previous section, we will now focus on the observable characteristics and detectable traces of deepfakes. Specifically, rather than redefining deepfakes, the aim here is to explain how different manipulation types produce spatial, temporal, frequency-domain, physiological, and forensic signals that can be used for detection and evaluation. More specifically, this distinction is important because a deepfake is not only defined by the fact that it is synthetic or manipulated, but also by the technical traces left by the generation, reenactment, rendering, and post-processing pipeline [55–57].

### *3.1. Deepfake Categories for Audio and Visual Objects*

While the previous section introduced a broad functional taxonomy based on modality, manipulation intent, and generation regime, this subsection focuses more narrowly on facial manipulation categories that are commonly used in deepfake detection research.

The two classifications function at different levels; hence, this distinction is important. The prior classification gives a broad conceptual framework of deepfakes in terms of modality, intent, and generation regime. The present subsection further translates the visual branch of that taxonomy into actual facial manipulation operations. In this instance, whole-face synthesis, identity swap, attribute manipulation, and expression swap become viable practical subtypes of visual deepfakes that often appear in face-forgery generation and detection studies.

Analytically, the recent literature proposes a practical classification of facial manipulations into four main categories, which is particularly helpful for both the analysis and mapping of detection research [56–58]:

1. Entire face synthesis: the creation of an entirely new face (or even an entire image/scene) that does not correspond to a real, recorded person.
2. Identity swap/face swap: replacing one subject's face with another's, typically while preserving the "host's" pose, lighting, and motion.
3. Attribute manipulation: changes to specific characteristics (age, gender/expression, morphological features), without necessarily changing the identity.
4. Expression swap/reenactment: we "transfer" the expressions/movements of a target from a source so that the mouth, eyebrows, and micro-expressions match another video or a live source. A classic (predating modern deep models but fundamental) family of reenactment techniques is exemplified in projects such as [29].

### *3.2. Generation-Side Artifacts and Observable Traces*

Deepfakes, even when visually convincing, often leave detectable traces. We see these traces because synthetic pipelines try to capture audiovisual information like a real camera, lens, sensor, microphone, light environment, and the human body in a physical sense. This may lead to inconsistencies in texture, geometry, motion, compression response, audio spectrum, and physiological plausibility levels.

Specifically, in this section, we pose an important question about how traces can be read. More specifically, they are assumed to be generation-side artifacts as signals that emerge from imperfect synthesis, reenactment, rendering, alignment, compression, or post-processing. Thus, this subsection aims to clarify why this kind of artifact appears in synthetic media. Lastly, it is worth noticing that we focused on showcasing how these observable traces are transformed into detection cues, or strategies for feature engineering, or inputs to the model or the decision signal for classifying deepfakes, localizing them, or recognizing their forgery type.

These traces are not "fixed" (they depend on the model, training quality, and post-processing), but they form a central basis for the logic of visual intelligence and digital forensics. These traces of operations are the following:

1. Spatial/morphological artifacts: Examples include imperfect blending at the face–skin/hair boundaries, unrealistic geometry in teeth/lips, inconsistencies in shading/lighting, or a "mismatched" background in high-frequency details. Such indicators are systematized in approaches that analyze visual artifacts as production "signatures" [59,60].
2. Temporal inconsistencies: In video, realism is strongly judged by frame-to-frame consistency: blink rate, facial micro-movements, lip-to-voice synchronization, head movement in relation to lighting/shading. A classic example of utilizing such a signal is the detection of unnatural blinking [61,62].

3. Frequency-domain traces: Many generative models leave traces in the frequency spectrum due to resampling and upsampling (e.g., "regularities" not systematically found in natural images). Frequency-domain analysis has been proposed as a complementary "channel" of evidence, especially when spatial cues are attenuated by compression [63,64].
4. Physiological cues: A more recent line of thinking capitalizes on the fact that real-time facial video incorporates subtle physiological changes (e.g., remote photoplethysmography based on micro-changes in skin color). Deviations in such signals can serve as an indication of synthetic origin [65,66].
5. Device/sensor-origin traces (forensic fingerprints): In digital forensics, camera model fingerprints help determine whether the content bears consistent traces of physical capture or has undergone alterations that destroy or replace them [67,68].

Lastly, alongside visual deepfakes, voice cloning/synthesis (audio deepfakes) has advanced rapidly. In research, evaluation datasets and protocols such as the one presented in [69] serve as "benchmarks" for detecting synthetic/spoofed speech and for comparing metrics across methods. In realistic scenarios, the threat is often multimodal: visual and audio signals may be simultaneously altered or "work together" to increase plausibility; this also affects how we define "deepfake characteristics" in practice [56].

## 4. Detection and Evaluation of Deepfakes: Methodologies, Data, and Benchmarks

Deepfake detection is a central subfield of multimedia forensics and content integrity. Unlike "closed" classification problems, deepfake detection is shaped by the continuous evolution of generative models, changes in distribution channels (platform compression, transcoding, screen recording), and the variety of operational scenarios (large-scale moderation versus forensic verification). In military use, the scenario may also involve rapid verification of battlefield media or command-related footage. As a result, evaluation cannot be limited to intra-dataset accuracy, but must systematically examine generalization, robustness, and decision calibration under "real-world" conditions [70–79].

As such, detection levels for what it means to detect a deepfake mean that three functional detection objectives are identified in the literature:

1. Authenticity classification (real vs. fake);
2. Spatial/temporal localization of alterations;
3. Classification of forgery types (e.g., swap, reenactment, synthesis).

This distinction is important because different datasets and benchmarks support different objectives: e.g., approaches such as [80] focus on detecting a forged face, while the other line of work extends the scope to a more "universal" analysis of forgery involving multiple tasks (classification and localization) [76–78].

### *4.1. Cues and "Signatures" of Forgery*

Detection methods utilize cues that can be grouped into the following feature-engineering levels:

1. **Spatial features:** Detectors convert local visual inconsistencies into measurable image features, such as texture irregularities, blending boundaries, facial warping, lighting mismatch, or abnormal local details. These features are commonly extracted from face crops, patches, or facial regions and are often used by convolutional neural network (CNN)-based and mesoscopic detectors [59,80–83].

2. **Temporal features:** Video detectors model the evolution of facial regions across consecutive frames. Instead of only observing that a video contains flickering or unnatural blinking, these methods measure frame-to-frame instability, motion inconsistency, lip-motion irregularity, or temporal feature drift using temporal aggregation, 3D CNNs, recurrent models, or transformers [61,62,84].
3. **Frequency-domain features:** Spectral detectors transform images or video frames into frequency representations and search for abnormal patterns introduced by upsampling, resampling, compression, or generative reconstruction. These features are especially useful when pixel-level artifacts are visually subtle or partly hidden by post-processing [63,64].
4. **Physiological and acquisition-based features:** Some detectors estimate biological or sensor-origin signals, such as rPPG, blink behavior, camera fingerprints, sensor noise, or compression history. These features are used to test whether the content remains consistent with natural human physiology or real camera acquisition [65,67,68].
5. **Multimodal consistency features:** In audiovisual deepfakes, detection may also examine whether the speech signal, speaker identity, lip motion, and facial dynamics agree with each other. This is important because a video may look plausible visually while still showing audiovisual synchronization or speaker-consistency errors [69,79].

As for physiological signal-based detection, however, it is worth noticing that it is sensitive to operating conditions. Specifically, the blood flow induces a subtle change in skin color that underpins remote photoplethysmography (rPPG) signals. As such, subtle color variations are weakened or corrupted due to compression and low resolution of the video, unstable lighting conditions, motion blur, occlusions, make-up, skin-tone variation, and platform re-encoding. This means that the face can be partly visible in real-world settings and recorded at a low frame rate. In addition, the camera auto-exposure and white-balance changes affect the reliability of physiological measurements. As a result, rPPG-based methods should be considered as additional forensic evidence rather than standalone proof of authenticity or manipulation. Evaluation should involve compressed, low-quality, and in-the-wild videos.

Lastly, multimedia forensics examines "device/camera traces," such as camera model fingerprints, as [67] proposes a CNN-based fingerprint that can be used as an indicator of consistency/inconsistency with natural capture.

### *4.2. Detector Families: From Frame-Level to Multimodal*

Based on how cues are utilized, detectors are broadly categorized as follows:

- Frame-level detectors (image-based detection): They focus on individual frames and offer advantages in terms of computational cost [85] and are used in large-scale screening, but they are vulnerable to the selection of "good" frames or to deepfakes that optimize spatial artifacts [86].
- Video-level detectors (spatio-temporal models): They incorporate temporal information (e.g., 3D CNN, temporal aggregation, transformers) and tend to improve detection in cases where the forgery "escapes" spatially but remains temporally inconsistent [84].
- Compact/mesoscopic architectures: for example, Ref. [83] presented a compact architecture for face-forgery detection, targeting meso-level features that remain useful under standard compression.
- Multimodal detection (audio–video): As deepfake production shifts toward full audiovisual synthesis, the integration of audio and video becomes particularly important. Datasets such as the ones used in [79] were designed specifically for evaluating multimodal scenarios (face + voice), while anti-spoofing benchmarks for speech support systematic evaluation of synthetic/transformed speech were also developed at times [69].

To conclude, when it comes to methodology, frame-level detectors learn spatial features from individual images while video-level detectors learn predictions on how facial region expressions and artifacts change on consecutive frames. The transformer-based and multimodal detectors, for instance, encode long-range temporal dependencies and also check cross-modal agreement, for example, whether mouthing agrees with the acoustic output.

To clearly show the detection methodologies, we propose a summary of the main families of deepfake detectors along with their technical principles, types of evidence, and their main strengths and weaknesses in Table 1 below. This taxonomy elucidates that detection is not an isolated technique, but rather several complementary approaches acting on a broad range of signals: From frame-level spatial artifacts to temporal, physiological, or even frequency-domain and multimodal consistency cues.

**Table 1.** Taxonomy of deepfake detection methods, technical principles, evidence types, strengths, and limitations.

| **Detection Family** | **Core Technical Principle** | **Main Evidence** | **Strengths** | **Main Limitations** | **Indicative References** |
|---|---|---|---|---|---|
| Frame-level/image-based detectors | Analyze individual frames as images and learn discriminative spatial features between real and manipulated content. | Texture inconsistencies, blending artifacts, facial boundary errors, lighting mismatch, abnormal local details. | Computationally efficient; useful for large-scale screening; can work even when only single images or extracted frames are available. | May ignore temporal inconsistencies; vulnerable when the fake is visually refined or when only high-quality frames are selected. | [59,80,85] |
| Video-level/spatio-temporal detectors | Model the evolution of visual features across consecutive frames using temporal aggregation, 3D CNNs, recurrent models, or transformers. | Frame-to-frame instability, inconsistent facial motion, unnatural eye movement, lip-movement irregularities, temporal flickering. | Better suited for video deepfakes; can detect manipulations that are not obvious in isolated frames. | Requires more computation and sufficient video length; performance may degrade after compression or frame-rate changes. | [61,62,84] |
| Frequency-domain detectors | Transform images or frames into the frequency domain and search for spectral anomalies introduced by generative upsampling, resampling, or compression. | High-frequency artifacts, spectral regularities, abnormal noise patterns, resampling traces. | Useful when spatial artifacts are visually subtle; complementary to pixel-level detection. | Sensitive to re-encoding, resizing, filtering, and platform transformations. | [63,64] |
| Physiological-signal detectors | Estimate biological signals from facial video and test whether they are consistent with natural human physiology. | Remote photoplethysmography, pulse-related skin color changes, blink behavior, | Can capture cues that generators often fail to reproduce naturally; useful for portrait videos. | Requires sufficient face visibility and video quality; may fail under poor lighting, low | [65,66] |
| | | micro-movement patterns. | | resolution, or heavy compression. | |
| Device/sensor forensic detectors | Examine whether the media contains traces consistent with real camera or sensor acquisition. | Camera fingerprints, sensor noise, compression history, metadata consistency. | Useful for forensic verification and chain-of-custody analysis. | May be weakened when metadata is removed or when the content is heavily processed. | [67,68] |
| Multimodal audio–video detectors | Compare information across modalities and detect inconsistencies between speech, facial motion, and speaker identity. | Audio–lip synchronization, voice–identity consistency, acoustic features, facial motion alignment. | Important for full audiovisual deepfakes; can detect cases where each modality alone appears plausible. | Requires synchronized audio and video; may be affected by dubbing, noisy audio, or legitimate editing. | [69,79] |
| Provenance-based verification | Verifies origin, editing history, and transformation chain through metadata, cryptographic manifests, signatures, or content credentials. | Content credentials, signed metadata, provenance chains, transformation logs. | Provides verifiable evidence beyond probabilistic detection; useful for institutional workflows. | Depends on adoption, metadata persistence, key management, and platform support. | [21,87–90] |
| Watermark-based verification | Embeds a detectable signal into generated or edited content so that synthetic media can later be identified or traced. | Embedded watermarks, watermark recovery signals, robustness under transformations. | Supports traceability and accountability at the generation stage. | Can be weakened by compression, cropping, re-encoding, fine-tuning, or removal attacks. | [91–94] |

### *4.3. Datasets and Benchmarks: What We Measure and Why It Matters*

Progress in detection depends largely on the available data. Datasets vary in various terms such as the types of forgery, levels of production "realism," abundance of sources/usage rights, compression/transformations rations, and evaluation objectives. As such, it is worth noticing the following datasets for benchmarking:

- DFDC Dataset (DeepFake Detection Challenge Dataset): large-scale data for benchmarking and systematic comparison of detection approaches [71,72].
- Celeb-DF: designed as a more challenging dataset to reduce the "ease" of detection via simplistic artifacts and push for generalization [73].
- DeeperForensics-1.0 focuses on conditions that closely resemble real-world pipelines and highlights the implications of cross-dataset evaluation [74].
- WildDeepfake: compiles an "in-the-wild" collection, highlighting the drop in performance when detectors are applied to real-world internet conditions [75].
- ForgeryNet and Challenge serve as a flexible benchmark for multiple forgery tasks (classification and localization) and as a challenge platform, enhancing the comparability of methods [76–78].
- FaceForensics++ serves as a classic evaluation dataset for manipulated facial images and is widely used in experimental comparisons [80].

Lastly, we notice that the different benchmarks differ not only in size but also in manipulation type, compression setting, realism, source diversity, and evaluation intention. Thus, the selection needs to be based on the specific objective of the study, essentially targeting any of the controlled benchmarking, cross-dataset generalization, in-the-wild robustness, multimodal detection, and forgery localization. To improve the usability of this survey, Table 2 provides a comparative summary of these datasets and benchmarks according to their modality, reported scale, manipulation type, availability, and intended evaluation use.

**Table 2.** Summary of major deepfake datasets and benchmarks.

| Dataset/Benchmark | Reported Scale | Modality | Manipulation Type | Availability/Access Route | Intended Use |
| --- | --- | --- | --- | --- | --- |
| DFDC | More than 100,000 videos; full version reported as 124 k videos | Video/ audiovisual | Face swapping using multiple manipulation methods | Public research dataset through official/project access routes | Large-scale training and benchmarking of deepfake detectors |
| Celeb-DF | 590 original celebrity videos and 5639 DeepFake videos | Video | High-quality celebrity face swapping | Public research dataset/project and mirror access routes | Challenging evaluation of face-swap detection and generalization |
| DeeperForensics-1.0 | 60,000 videos and 17.6 million frames | Video | End-to-end face swapping with real-world perturbations | Research dataset/project access | Robustness testing under realistic visual degradation |
| WildDeepfake | 7314 face sequences from 707 internet deepfake videos | Video/face sequences | In-the-wild deepfake videos collected from online sources | Public research dataset/GitHub access | Testing detector performance under real-world internet conditions |
| ForgeryNet/ForgeryNet Challenge | 2.9 million images and 221,247 videos | Image and video | Multiple image-level and video-level forgery methods | Research benchmark/ challenge access | Classification, spatial localization, video forgery detection, and temporal localization |
| FaceForensics++ | More than 1.8 million manipulated images from over 1000 videos | Video/frame-level images | DeepFakes, Face2Face, FaceSwap, NeuralTextures | Public benchmark dataset | Standardized evaluation of face manipulation detection under compression settings |

### *4.4. Evaluation Protocols and Metrics: From AUROC to Operational Reliability*

The literature agrees that evaluation must be multi-level:

1. Intra-dataset vs. cross-dataset evaluation: Cross-dataset performance better captures domain shift and is closer to real-world conditions, where the generator or compression channel is not known in advance [73–75,85].
    - Intra-dataset evaluation determines whether a detector can detect similar patterns to those seen during training, while cross-dataset evaluation tests whether the learned indications are still valid when the manipulation method, source camera, level of compression, or distribution platform changes. Because of this distinction, many detectors learn dataset-specific artifacts instead of general forensic evidence of manipulation.
2. Classification metrics and class imbalance: In addition to accuracy, AUROC and Area Under the Precision-Recall Curve (AUPRC) are used (especially in cases of class imbalance), while in anti-spoofing scenarios, metrics such as Equal Error Rate (EER) are standard [69].

- While these metrics technically measure different things, they reflect the detector's behavior. Area Under the Receiver Operating Characteristic Curve (AUROC) measures the ability to distinguish between real and fake samples over thresholds. AUPRC is informative when the fake content is rare. EER is the operating point at which false acceptance and false rejection are equal. A detector that performs well in tests may be of limited operational use if the threshold has been poorly calibrated.

3. Robustness tests: Tests under re-encoding, resolution changes, cropping, filtering, and platform transformations are considered essential, as these steps often "neutralize" surface artifacts [71–75,79].
    - From a more technical standpoint, these transformations change the pixel, frequency, and compression characteristics of the media. As a result, they can weaken or even remove the artifacts that the detector uses for classification. In other words, robustness testing evaluates whether the detector is solely dependent on fragile surface signatures or if it can maintain effectiveness in realistic degradation and redistribution.
4. Calibration and decision thresholds: In operational scenarios, the output is not merely a "label," but a risk score that leads to action (e.g., human review, takedown, posting ban). Therefore, calibration and threshold selection are part of the evaluation.
    - Calibration refers to the notion of whether the confidence score of a detector corresponds to the actual probability of a manipulation instead of a high or low number. The next step determines the evidence needed to trigger an action. It is particularly important in high-stakes situations where false positives can harm trust in authentic content and false negatives can allow harmful synthetic media to circulate.

Also, it is worth mentioning that the detection of deepfakes is evolving into an arms race. For this reason, complementary strategies are being developed that shift the focus from "assessing the likelihood of forgery" to "provenance." In this context, the Coalition for Content Provenance and Authenticity (C2PA) specification proposes a technical framework for statements of provenance and content integrity [87], while organizations such as the Content Authenticity Initiative (CAI) promote interoperability and adoption practices [88]. The combined approach (detection + provenance) aims to reduce both false positives and the "strategic doubt" that arises when society is unable to distinguish between genuine and synthetic content [70,87,88].

Lastly, another important challenge is that deepfake detectors themselves can be subjected to an adversarial attack. Under this setting, the attacker not only tries to generate realistic synthetic media but also modifies the output so that it evades detection. As a result, these attacks can include slight visual distortions or modifications (which are aware of the encoding), special post-processing of the detector, or adaptive synthesis, optimizing the synthetic content against known forensic cues.

Based on the above, this creates a moving-target problem: a detector trained on current artifacts may perform well in controlled evaluation, but may fail when attackers deliberately suppress, distort, or replace the signals on which the detector relies. Given this, future evaluation protocols should include a variety of adaptive and adversarial robustness tests, not just accuracy on standard datasets or cross-dataset generalization [19,86].

### *4.5. Recent Detector Models and Reported Benchmark Performance*

Table 3 provides an indicative summary of representative recent detector models and their reported benchmark performance. The values are presented as approximate ranges or representative scores, since results vary across dataset versions, compression levels, preprocessing pipelines, train/test splits, and evaluation protocols.

**Table 3.** Indicative summary of recent deepfake detection models and reported benchmark performance.

| Model/Approach | Model Family | Benchmark Dataset(s) | Approximate Reported Performance | Evaluation Context |
|---|---|---|---|---|
| DFDT | Vision Transformer-based detector | FaceForensics++, Celeb-DF, WildDeepfake | Around 99% AUC on controlled FaceForensics++/Celeb-DF settings; around 80–81% on WildDeepfake | Frame-level benchmark evaluation |
| SBI | Self-blended image training | FF++, Celeb-DF, DFDC, DFDCP | Generally improves cross-dataset AUC by several percentage points compared with common baselines | Generalization to unseen manipulations |
| FSBI | Frequency-enhanced self-blended images | FF++/Celeb-DF | Around mid-90% AUC on Celeb-DF under cross-dataset evaluation | Frequency-domain and cross-dataset robustness |
| CNN/Transformer comparative models | CNN and transformer architectures | FF++, Google DFD, Celeb-DF, DeeperForensics, DFDC | Often above 90% AUC in controlled intra-dataset settings, but lower under cross-dataset testing | Comparison of CNN-based and transformer-based detectors |
| Deepfake-Eval-2024 evaluated detectors | Open-source and commercial detectors | In-the-wild multimodal 2024 benchmark | Substantial performance drops compared with older academic benchmarks; reported AUC decreases of roughly 45–50% in some modalities | Real-world, multimodal, in-the-wild robustness evaluation |

## 5. Mitigation, Governance, and Regulatory Compliance for Synthetic Content

Building upon the defense-in-depth model introduced in Section 1, modern mitigation strategies treat forensic detection as merely one subsystem. The necessity of this integrated architecture, combining detection with provenance and governance, stems from two consistent findings in the literature:

- A detector's performance on controlled data does not guarantee operational reliability under varying dissemination channels, and
- The "trade-off" between generation and detection is dynamic, thus requiring continuous evaluation and revision of controls [95,96].

### *5.1. Threat Identification, Risk Modeling, and Transparency Disclosure Controls*

Threat modeling can classify deepfake-related harms into several risk categories, including information integrity risk, identity integrity risk, personal harm or harassment, and evidentiary risk. As such, this taxonomy allows for the mapping of controls at each stage of the lifecycle (create–edit–publish–moderate–archive), as well as the definition of auditability and traceability requirements (e.g., who viewed, who evaluated, who decided, based on what evidence) [95,97].

This risk-based classification also provides the basis for deciding when transparency and disclosure measures should be activated as operational controls. As such, at the regulatory level, there is a growing emphasis on requirements mandating disclosure that content is artificially generated or modified when it has the potential to be misleading (i.e., with deceptive potential). Such transparency obligations serve as a "first line of defense" because they reduce the scope for deception regardless of the performance of detectors [98]. For practical implementation, the disclosure must be: salient, persistent upon redistribution, and unambiguous regarding what it denotes (i.e., in terms of semantic clarity: "AI-generated," "AI-edited," "composite," etc.), to avoid false impressions or over-labeling that lead to "trust fatigue".

### *5.2. Risk Governance, Management Systems, and Technical Content Transparency*

For compliance to be operationally feasible, a structured management system with clear roles, procedures, and documentation is required. International Organization for Standardization/International Electrotechnical Commission (ISO/IEC) 42001:2023 AI Management System (AIMS) supports the establishment of policies, objectives, and controls for organizations that develop or use AI systems, while ISO/IEC 23894:2023 guides risk management specifically in the context of AI systems [99,100].

In the field of deepfakes, the implementation of these frameworks involves:

- Testing, evaluation, verification, and validation (TEVV) (testing–evaluation–verification–validation) for detection/labeling/provenance tools;
- Assurance case (documented argumentation that the system is "sufficiently secure/reliable" for a specific use);
- Change management (what changes when the codec, model provider, or platform changes);
- Monitoring and drift management (monitoring of performance degradation/increase in errors).

The essence is that the technical solution is to be transformed into a controlled system with measurable requirements and accountability mechanisms.

Within this governance structure, technical content transparency functions emerge as the evidence layer that supports verification, accountability, and auditability. As such, technical transparency (content transparency) includes approaches such as metadata-based provenance, watermarking, and tamper-evident mechanisms. In systems based on provenance, the technical principle is to attach verifiable metadata or cryptographic manifests to media files so that subsequent users can inspect their origin, editing history, and chain of modifications. By contrast, watermarking embeds a signal into the content (the image) itself, meaning it is clear this has been embedded and is detectable. Whether it can be detected after compression, cropping, re-encoding, or a deliberate attempt to remove it is a question of robustness. One of the main concerns one must take into consideration is the distinction between:

- Detection signals (probabilistic evidence from Machine Learning detectors);
- Verifiable evidence that can be verified cryptographically or through structured manifests.

Furthermore, according to experts in the industry and forensics, a system must have trust and verification pipelines for capture time. Specifically, this is important as it will make it capable of generating trustworthy pieces of digital evidence at the layer. More specifically, capture-time verification is a novel hardware-assisted technology built into the device recording the content (e.g., a secure hardware enclave or lens cryptographic camera), rather than an after-the-fact deepfake detector which tries to detect manipulation post-event. This is important as, by using cryptographic signatures on the media and related metadata (time, location, device ID) at the time of capture, it creates an unchangeable root of trust in these systems. As a result, the system is then able to indicate that any further creation, generative alteration, or illicit editing breaks that tamper-proof hardware chain of custody so that it can be irrefutably proven that this file has been changed from its true original condition.

National Institute of Standards and Technology (NIST) AI 100-4 ([95]) summarizes technical guidelines for digital content transparency, emphasizing that no single technique is sufficient on its own; value arises from clear threat models, resilience testing, and integration into processes [86]. In practice, provenance logic also requires key management (key/signature management), key revocation/rotation policies, and mechanisms for "handling provenance gaps" (e.g., legacy material or metadata loss).

### *5.3. Watermarking Resilience, Operational Integration, and Forensics-Grade Documentation*

Watermarking addresses the problem of persistent marking under channel distortions: compression, recoding, scaling, cropping, and redistribution. The Advanced Television Systems Committee (ATSC) standards for audio/video watermark embedding and for content recovery in redistribution reflect mature technical logic: clear specifications for signal embedding, recovery conditions, and operational scenarios under channel noise [101–103]. For synthetic media, the analogy is useful: a watermark is only valuable if it is accompanied by measurable robustness guarantees, a clear definition of failure modes, and evaluation procedures against removal/tampering attacks.

These robustness requirements connect watermarking to operational and forensic workflows, where technical signals must be documented, interpreted, and validated within controlled decision processes. As a result, a comprehensive approach to deepfakes requires linking detection techniques with decision workflows: when to perform automatic flagging, when human-in-the-loop intervention is required, how to define thresholds, and how to document the decision. In high-stakes environments (e.g., evidentiary use), chain-of-custody procedures, transformation logging, and reproducibility of the evaluation methodology are essential. Initiatives to evaluate analytical systems for AI-generated deepfakes reinforce precisely this dimension: standardized tests, clear limits on conclusions, and a distinction between "what a tool proves" versus "what it assumes" [96].

At the same time, the ethical frameworks of the United Nations Educational, Scientific and Cultural Organization (UNESCO) and Organisation for Economic Co-operation and Development (OECD) serve as a horizontal design principle: proportionality of measures, avoidance of disproportionate side effects, accountability, and appeal/redress mechanisms, so that technical enforcement does not become a source of injustice [104,105]. Overall, effectiveness is judged as a system property: disclosure/transparency [98], risk governance [99,100], technical transparency [95], and forensics-grade documentation [96], with a legal understanding of the risks to democracy/security/privacy [97].

## 6. Implications and Scenarios of Abuse: A Case-Based Analysis in Key Application Areas

Deepfakes and, more broadly, synthetic media pose a challenge that transcends the "true/false" dichotomy. The critical dimension is the reconfiguration of trust mechanisms: which narrative is deemed credible, how is accountability assigned, how is the authenticity of an audiovisual document verified, and what is the organizational cost of verification. The international literature points out that the consequences are systemic, as they are linked to the speed of dissemination, personalization/targeting, and the potential for escalating disinformation at low production costs [106–110]. At the same time, the impacts vary significantly by sector, depending on whether the primary concern is the public sphere (information integrity), identity verification (identity proofing), forensics/chain of custody, or the protection of vulnerable groups (targeting, non-consensual material) [106–111].

To make the discussion more concrete and easier to follow regarding the subsections below, we note that we examine four representative application areas. Firstly, Case A focuses on news, journalism, and fact-checking, where the main concern is the erosion of evidentiary trust and the verification of public information. Secondly, Case B examines financial fraud, corporate security, and remote identity verification, where deepfakes are used primarily for impersonation and to influence high-value operational decisions. Thirdly, Case C addresses public-sector, law enforcement, and forensic contexts, where the reliability of audiovisual evidence and chain-of-custody procedures are central. Lastly, Case D focuses on education and student protection, where synthetic media can affect academic integrity, learner safety, and incident-response practices.

### *6.1. Case A—News, Journalism, and Fact-Checking (Newsroom Verification)*

In the news environment, deepfakes act as a catalyst for "evidentiary erosion": the increased likelihood of synthetic content amplifies uncertainty and enables the strategic denial of authentic evidence (liar's dividend), thereby undermining accountability [109,110]. In practice, verification in a newsroom is not a purely technical assessment (detector score), but a combination of several factors such as dissemination and source analysis (first post, reproduction networks, signs of coordination), source/metadata verification (where available), and multimodal consistency (audio–lip-sync–lighting–temporal coherence) and cross-referencing with independent contexts [89,111].

In this context, provenance approaches (e.g., Content Credentials/C2PA) are viewed as a complementary "line of defense" against exclusive reliance on detection, as they shift the focus from "does it look fake" to "what provenance history does the file carry" [89,90]. Lastly, the European trend toward transparency requirements for certain uses/outputs of digital content reinforces the value of documented labeling and disclosure procedures, particularly when the content has the potential to be misleading [112,113].

### *6.2. Case B—Financial Fraud, Corporate Security, and Remote Identity Verification (Know Your Customer/Remote Onboarding)*

In financial and corporate settings, the threat shifts from the "public narrative" to operational identity fraud (impersonation fraud), where voice cloning, synthetic video calls, and biometric abuse can influence high-value decisions (e.g., payment orders, data changes, remote registration) [106–108,114]. The technical challenge is not limited to detection but rather includes the design of risk-based identity proofing and "adaptive friction" for critical actions: combining biometric and non-biometric signals, documenting controls, and establishing clear out-of-band verification channels for high-risk decisions [114,115].

From a compliance perspective, the processing of biometric data may be subject to heightened data protection requirements; therefore, organizational maturity requires documentation of the legal basis, purpose, proportionality, and security measures [116]. The literature notes that in these scenarios, resilience is achieved primarily through a combination of procedural controls and technical measures, not through a single detection tool [106–108,114].

### *6.3. Case C—Public Sector, Law Enforcement, and Forensic Use (Forensics, Chain of Custody)*

In law enforcement and the justice system, deepfakes affect both the production of falsified evidence and the reliability of authentic records. Specific analyses for law enforcement emphasize that the response must be based on standardized chain-of-custody procedures, secure collection and storage, recording of transformations, and traceability of the verification process [117].

In this context, technical detection is part of a broader "evidence package" that includes documented collection, hashing, logging, assessment of method limitations, and, where possible, provenance verification. This need is reinforced by the fact that modern synthesis models and platform-specific transformations (re-encoding, downsampling) can alter or eliminate traces, requiring caution in drawing conclusions and ensuring the reproducibility of the process [95,117].

### *6.4. Case D—Education, Academic Integrity, and the Protection of Students*

In the educational setting, the implications unfold along two main lines. Firstly, the production of synthetic "evidence" (images/videos/audio) can undermine academic assessment and increase the scope of deception beyond traditional plagiarism [108,111]. Secondly, targeting with deepfakes (particularly non-consensual synthetic material and defamatory videos) has serious implications for the safety and psychosocial well-being of learners [107,108].

At the level of policy and pedagogical practice, the literature supports the integration of "evidence hygiene" procedures: mandatory documentation of material sources in assignments, the development of media literacy skills, and clear incident management workflows (reporting–recording–support–removal of reposts) [108,118]. The UNESCO ethical framework highlights the need for a human-centered approach, particularly when surveillance/detection measures may produce disproportionate side effects [118].

### *6.5. Synthetic Assessment: Effectiveness as a System Property*

A comparative analysis of the above cases reinforces a consistent conclusion: countermeasures are effective only when approached as a holistic system property. The international literature confirms that integrating detection, provenance, and organizational risk management is essential for operational resilience across diverse sectors [90,95].

In the European context, transparency obligations and accountability/due diligence requirements for platforms act as a catalyst for the institutional integration of these practices [112,113]. Similarly, guidelines on digital identity and risk-based proofing propose combinations of controls and adaptive measures that mitigate harm even when detection is not "perfect" [114,115].

## 7. Conclusions

The current literature shows that deepfakes have moved from an isolated forgery technique to a wider synthetic content ecosystem, where creation, dissemination, and persuasiveness interact with platforms, social networks, and real-world information inequalities [70,107]. The next phase of this domain is expected to be shaped by higher video and audio fidelity, stronger temporal consistency, and near-real-time operation in avatars, teleconferences, customer service systems, and live-style broadcasts. At the same time, the convergence of image, audio, and text generation means that deepfakes are no longer only visual imitations. They can now become complete synthetic personas, with convincing speech, language, facial expression, and paralinguistic behavior. This creates serious risks for corporate fraud, political disinformation, identity abuse, and targeted harm against individuals. However, the same technologies can also support legitimate uses in film production, accessibility, education, and creative media, which makes it necessary to assess deepfakes according to use, context, intent, and potential harm [106,107].

The threat landscape is also changing as attacks are moving away from single fake videos or single-medium manipulations and toward more complete fraud workflows. These may include target data collection, voice cloning, face-swapped video production, fake documents or profiles, and social engineering. This means that the effectiveness of a deepfake does not depend only on technical quality. It also depends on the social context, the trust relationship between sender and receiver, the distribution channel, and existing inequalities or vulnerabilities [70]. In education and public communication, repeated exposure to synthetic content can increase source confusion and make it harder for users to distinguish reliable from unreliable information. For this reason, media literacy and information literacy are not secondary issues; they are part of the wider response to deepfakes [108,111].

From a technical standpoint, the main challenge is no longer only to improve detection accuracy on known datasets. The more important challenge is operational reliability under domain shift, unknown generators, new compression methods, different distribution channels, and multimodal forgeries that combine face and voice. Detection systems therefore need to be evaluated not only in controlled research settings, but also under realistic conditions. This requires unified and reproducible evaluation protocols across multiple datasets, with clear reporting of false positives, false negatives, calibration, robustness, and practical trade-offs. Benchmarks and evaluation tools such as DeepfakeBench are important because they bring together multiple datasets, protocols, and detection methods, helping to connect research performance with operational deployment [119]. At the same time, systematic surveys remain necessary because they organize the field by creation methods, detection categories, and open research gaps, including robustness, interpretability, and calibration [120].

As demonstrated, standalone detection is vulnerable to distribution shifts and novel generation methods, necessitating the integrated defense architecture discussed throughout this survey. The standards, such as the C2PA family of specifications, are important because they provide a technical way to document the origin and transformation history of digital content, including generative content [21,90]. At the organizational level, standards such as ISO/IEC 42001 for AI management systems and ISO/IEC 23894 for AI risk management guidance provide a useful framework for policies, controls, auditing, lifecycle documentation, and continuous improvement [99,100]. In practice, this means that content integrity should be treated as a process, not only as a detector output.

Watermarking is another important part of this wider strategy, especially for diffusion and generative models. Recent approaches move watermarking closer to the production stage, so that generated content carries a detectable signature without obvious visible distortion. Stable Signature, for example, embeds a watermark into latent diffusion models by adapting the decoder, with the aim of supporting robust signature recovery after common transformations [91]. Tree-Ring Watermarking uses a fingerprint in the initial noise vector and structures it in Fourier space, aiming to remain detectable after transformations such as cropping or rotation [92]. These methods show that watermarking can support tracking and accountability. However, watermarking should not be treated as a complete solution. The literature also shows that watermarks can be weakened, removed, or attacked through fine-tuning and other techniques [93,94]. Therefore, watermarking needs continuous evaluation, clear robustness guarantees, and integration with provenance standards, platform-level measures, and governance processes [91–95].

Overall, the most realistic response to deepfakes is a multi-level strategy, as regulation is also becoming more mature. In the European context, the trend is moving from general recommendations toward clearer transparency, accountability, and due diligence obligations. The Digital Services Act requires risk management and transparency measures, especially for very large online platforms, with the aim of reducing systemic risks linked to misinformation and harmful content distribution [23,113]. However, legal rules alone cannot solve the problem. Work on public discourse and democratic resilience shows that deepfake governance must combine technical tools, platform enforcement, victim support and redress mechanisms, institutional safeguards, and systematic media literacy. As such, it is worth noticing that future work should focus on evaluation under real-world conditions, robustness against adaptive attacks, clearer documentation of failure modes, and stronger links between technical standards, governance frameworks, and social resilience.

In summary, the defense against deepfakes rests on three operational pillars. First, detection provides probabilistic forensic evidence by identifying spatial, temporal, frequency-domain, physiological, and multimodal inconsistencies. Second, provenance strengthens authenticity by documenting the origin, transformation history, and integrity of digital content through metadata, signatures, content credentials, and watermarking. Third, governance ensures that technical measures sit within transparent, accountable, legally compliant, and human-supervised processes. Taken together, these three pillars show that deepfake resilience cannot rely on a single detector or platform rule; it requires an integrated socio-technical defense architecture.

**Author Contributions:** Conceptualization, A.G., N.E.M. and S.P.; methodology, A.G.; software, A.G.; validation, A.G., E.K., T.V., N.E.M. and S.P.; formal analysis, A.G.; investigation, A.G.; resources, A.G., E.K., K.S. and T.V.; data curation, A.G.; writing—original draft preparation, A.G. and S.P.; writing—review and editing, A.G., E.K., K.S., T.V., N.E.M. and S.P.; visualization, A.G.; supervision, N.E.M. and S.P.; project administration, A.G.; S.P. and A.G. carried out the main workload of the manuscript, including the literature search, testing, analysis, reference collection, drafting, and revisions. E.K. contributed to validation, reference identification, reviewing, and editing. K.S. reviewed the manuscript, added missing references, and improved recent sections and tables. T.V. supported rewriting, validation, testing, reference checking, and the development of additional ideas. N.E.M. provided supervision, validation, editing, conceptual input, and critical feedback. S.P. supervised the work, co-developed the original draft with A.G., and contributed to major revisions. All authors have read and agreed to the published version of the manuscript.

**Funding:** This research received no external funding.

**Institutional Review Board Statement:** Not applicable.

**Informed Consent Statement:** Not applicable.

**Data Availability Statement:** No new data were created or analyzed in this study. Data sharing is not applicable to this article.

**Acknowledgments:** During the preparation of this manuscript/study, the authors used DeepL Translator, version 26.2.1, developed by DeepL SE, Cologne, Germany, as a neural machine translation tool for the purposes of Greek-to-English translation and linguistic refinement of author-prepared text. Also, during the preparation of this manuscript/study, the authors used Grammarly, version 1.2.210.1786, developed by Grammarly, Inc., for the purposes of English-language editing, grammar checking, spelling correction, and readability refinement of the English manuscript version. Lastly, during the preparation of this manuscript/study, the authors used ChatGPT, GPT-5.3, developed by OpenAI, a tool for the purposes of limited language polishing, structuring, formatting support, and checking Grammarly-assisted outputs for awkward English wording, passive-to-active voice suggestions, punctuation issues, grammar errors, and minor inconsistencies in flow. Specifically, for the GPT-5.3 model's input: for example, the authors used prompts such as: "Check the text for inconsistencies in flow, grammar mistakes, or unnatural English wording, and suggest how to rewrite it while changing as little as possible from the original text and explaining the rationale of your suggestions." Sections: Abstract, 1, 2.2, 2.3, 4.2, 4.4, 5.3, and 7. These tools were not used to generate scientific content, methodology, results, data analysis, interpretations, or conclusions. All translated and edited texts were reviewed, verified, and approved by the authors. The authors have reviewed and edited the output and take full responsibility for the content of this publication.

**Conflicts of Interest:** The authors declare no conflicts of interest. Author Kleanthi Santamouri is the manager of Salvezza Energy Systems Ltd. This affiliation did not influence the conception, methodology, analysis, interpretation, writing, review, or conclusions of the manuscript.

**DURC Statement:** The current research is limited to the academic and educational study of deepfakes, synthetic media, multimedia forensics, content provenance, and AI governance. It contributes to the responsible understanding of deepfake generation, detection, verification, transparency, and regulatory compliance, with the primary aim of supporting content integrity, public awareness, ethical digital media practices, and responsible AI education. This research is review-based and does not provide operational procedures, source code, datasets, technical instructions, or deployment guidance that could enable harmful use. Therefore, it does not pose a threat to public health or national security. The authors acknowledge the dual-use potential of research (DURC) involving deepfakes, synthetic media generation, and content verification, particularly in relation to misinformation, identity impersonation, evidence fabrication, and security-sensitive communication contexts. The authors confirm that all necessary precautions have been taken to prevent potential misuse. As an ethical responsibility, the authors strictly adhere to relevant national and international laws concerning dual-use research of concern, responsible AI, data protection, and digital content integrity. The authors advocate for responsible deployment, ethical considerations, regulatory compliance, transparent reporting, provenance-based verification, and human oversight to mitigate misuse risks and foster beneficial outcomes.

## Abbreviations

The following abbreviations are used in this manuscript:

| | |
|---|---|
| AI | Artificial Intelligence |
| AIMS | AI Management System |
| ATSC | Advanced Television Systems Committee |
| AUPRC | Area Under the Precision-Recall Curve |
| AUROC | Area Under the Receiver Operating Characteristic Curve |
| C2PA | Coalition for Content Provenance and Authenticity |
| CAI | Content Authenticity Initiative |
| CNN | Convolutional Neural Network |
| DDIM | Denoising Diffusion Implicit Model |
| DDPM | Denoising Diffusion Probabilistic Model |
| DFDC | DeepFake Detection Challenge |
| DSA | Digital Services Act |
| DURC | Dual Use Research of Concern |
| EER | Equal Error Rate |
| EU | European Union |
| GAN | Generative Adversarial Network |
| GDPR | General Data Protection Regulation |
| GPT | Generative Pre-trained Transformer |
| ISO/IEC | International Organization for Standardization / International Electrotechnical Commission |
| NIST | National Institute of Standards and Technology |
| OECD | Organisation for Economic Co-operation and Development |
| RGB | Red–Green–Blue |
| rPPG | Remote Photoplethysmography |
| TEVV | Testing, Evaluation, Verification, and Validation |
| UNESCO | United Nations Educational, Scientific and Cultural Organization |